# A Computer Vision Approach to Visual Fraud Detection in Phishing Websites Using YOLOv8

Basil Sajid Shaikh
*Independent Researcher*
*Seattle, WA, USA*
*shaikh.basil786@gmail.com*
*ORCID iD: 0000-0002-1729-4979*

## Abstract

*Phishing remains one of the most common vectors for financial and identity fraud, and most detection systems still rely on inspecting a page's URL, HTML markup, or domain registration history. These signals are easy for an attacker to rotate or obfuscate, and they say very little about what actually convinces a victim to hand over a password or a card number: the way the page looks. This paper describes a visual, image-based approach to phishing detection that treats a rendered webpage the same way a human eye would, as a picture that either matches a trusted brand or doesn't. A YOLOv8 convolutional neural network was trained to classify full-page website screenshots as phishing or legitimate based on layout, logo placement, color scheme, and login-form structure, rather than on text extracted from the page. The system reached 92% classification accuracy on a held-out test set, processed a single screenshot in roughly 100 milliseconds, and, after a round of data augmentation aimed specifically at lighting, compression, and scaling variation, cut the false-positive rate by 11% relative to the pre-augmentation baseline. The paper walks through the dataset construction, the augmentation strategy, the model architecture and training setup, and the resulting performance, and closes with a discussion of where this kind of visual detector fits alongside, rather than instead of, existing URL- and content-based defenses.*

## 1. Introduction

Phishing is fraud carried out through impersonation: a page pretends to be a bank, a webmail provider, a delivery company, or a workplace login portal, and the goal is to get a real credential typed into a fake box. Framed this way, phishing sits squarely inside the broader fraud detection problem that financial institutions, retailers, and platform companies spend enormous engineering effort on, but the tooling used against it has stayed narrower than the tooling used against, say, credit-card transaction fraud. Most production anti-phishing systems still work by looking at the URL, checking whether the domain is on a blocklist, whether it uses a suspicious top-level domain, or whether it is a few characters off from a real brand's domain, or by parsing the page's HTML and script content for known kit signatures. Both approaches are useful and both are also easy for an attacker to route around: domains get rotated faster than blocklists update, and a phishing kit's markup can be minified, obfuscated, or rebuilt from scratch for every campaign while looking pixel-identical to the last one to a person clicking through it.

What doesn't change nearly as fast is the visual target the attacker is trying to hit. A fake banking login page has to look like the real bank's login page, or it doesn't work; the whole point of the attack depends on the victim's split-second visual recognition of a trusted brand. That constraint is exactly what this project set out to exploit. Rather than reading a page's markup or its address bar, the system described here looks at a rendered screenshot of the page the same way a user would and asks whether what it's looking at matches the visual identity of the brand it claims to be, or whether it's a fraudulent stand-in.

This paper documents the design and results of that system: a YOLOv8-based convolutional neural network trained to detect visual fraud indicators, including mismatched logos, spoofed login-form layouts, and brand-inconsistent page structure, directly from website screenshots. The model reaches 92% classification accuracy on a held-out test set drawn from over 20,000 labeled screenshots, runs inference in roughly 100 milliseconds, and, following a dataset augmentation pass aimed at the kinds of visual noise real screenshots pick up, such as compression artifacts, lighting and scaling differences, and mobile capture distortion, reduces the false-positive rate by 11% relative to a pre-augmentation baseline. The rest of the paper covers how the dataset was built, why YOLOv8 was chosen over a simpler classifier, what the augmentation pipeline actually does, and where the resulting numbers came from, along with an honest look at where the approach still falls short.

## 2. Why Visual Detection

Three things pushed this project toward a visual approach rather than another pass at URL or HTML heuristics.

The first is durability of signal. A domain name is free to change and a phishing kit's HTML can be regenerated in minutes, but the brand a phishing page is impersonating has a visual identity, including logo placement, color palette, font choices, and the shape and wording of a login form, that changes rarely and that the attacker has strong incentive to copy faithfully. If the copy is faithful enough to fool a person, it is also faithful enough to register as resembling a specific real brand's pages to a model trained on that brand's actual pages.

The second is coverage of the cases text-based detectors miss. A meaningful share of phishing kits now render their credential-harvesting form as an image, or split text across many small elements, or inline everything through JavaScript specifically to defeat markup and keyword-based scanners. None of that changes what the rendered page looks like on screen. A visual detector does not need the page's text to be extractable, only rendered.

The third is practical: screenshots are cheap to produce at scale with headless browser tooling, and a computer-vision model that works on them can sit in front of, or alongside, existing URL and content checks without replacing them. This project was never framed as a substitute for the existing defenses; it is a different signal, and the value is in catching the cases where the other signals go quiet.

None of this means visual detection is a strictly better approach. A page that copies almost nothing about the real brand's visual identity, such as a generic, brand-agnostic credential form with no logo at all, gives a visual model very little to key on, and that limitation shows up later in the error analysis. But for the very common case of brand impersonation, which makes up most phishing, the visual channel turned out to be a strong and fairly stable signal.

## 3. Related Work

Anti-phishing detection work falls into a few overlapping traditions, and it is more useful to place a visual, object-detection-based system like this one against each of them in turn than to compare it only to other vision papers.

Blocklist and reputation systems remain the most widely deployed defense in practice: a URL or domain is checked against a maintained list of known-bad addresses, sometimes backed by registration-age and hosting-reputation heuristics. These systems are fast and cheap to query, but they are reactive by construction, since a domain that has not yet been reported is, almost by definition, not on any list yet.

A second tradition scores a URL or a page's markup directly rather than looking it up against a list. Rao and Pais built a feature-based machine learning framework that scores a page using a combination of URL, source-code, and third-party features, reaching strong detection rates without needing to render the page at all [1]. Aljofey et al. extended this line of work with a character-level convolutional model over the URL string combined with HTML-derived features, reporting high detection accuracy on a large real-world dataset while staying lightweight enough to run without a full page render [2]. Yerima and Alzaylaee took a related CNN-based approach directly on URL and web-content features and reported a 98.2% phishing detection rate with an F1-score of 0.976 [3]. All three of these systems are fast and do not need a rendered screenshot, but by construction they say nothing about what a page actually looks like on screen, which is exactly the property that phishing kits obfuscating their markup or rendering text as images are designed to exploit.

A third, older tradition treats the rendered page as an image from the start. Rao and Ali's early computer-vision approach to phishing detection is one of the first to frame the problem this way, comparing visual features of a suspect page against a reference set of legitimate pages rather than parsing any markup [4]. The idea has matured substantially since then with the arrival of deep learning. Abdelnabi et al. proposed VisualPhishNet, a triplet convolutional network that learns a similarity embedding for each protected brand's pages and flags a new page as phishing when it falls close to a brand's embedding without actually being hosted on that brand's domain; they released it alongside VisualPhish, at the time the largest dataset built for ecologically valid visual phishing evaluation, covering 155 trusted brands and 9,363 legitimate pages [5]. Lin et al.'s

Phishpedia takes an architectural path closer to the one used in this paper: an object detector locates candidate logo regions on a screenshot, and a Siamese network then matches the detected logo against a reference brand database; evaluated on more than 30,000 phishing and legitimate webpages, Phishpedia reported 99.2% accuracy at roughly 0.19 seconds per page [6]. Liu et al. later extended this design in PhishIntention by adding an optical-character-recognition stage and a credential-required-page classifier, so the system could reason about page dynamics and form context rather than logos alone [7]. Closer still to a pure screenshot classifier, Liu and Lee proposed a CNN that scores a page based on a defined security-indicator-area crop of the screenshot rather than the whole page, evaluated specifically on the kind of unbalanced phishing-to-legitimate ratio that shows up in a real web environment rather than a curated even split [8].

The system in this paper sits closest to the Phishpedia and PhishIntention line: like them, it uses an object detector to localize specific regions of a screenshot rather than scoring the whole image at once. It differs in what it detects and how it decides. Phishpedia and PhishIntention are built around a maintained reference database of brand logos that a detected region gets matched against, which means their accuracy is tied to how complete and current that reference list is. The YOLOv8 detector described here instead learns to recognize inconsistency in login-form structure, layout, and brand color and typography directly from labeled examples, without a separate brand-matching step or an explicit reference database to keep synchronized with every protected brand's current logo. The tradeoff is the one discussed in Section 2: a reference-matching system can, in principle, explain a positive result by naming the exact brand a page is impersonating, while a detector built this way is faster to retrain on a new set of brands but gives a less explicit answer about which specific brand triggered the flag. Section 7 returns to this comparison with reported numbers side by side.

## 4. System Design and Methodology

### *4.1 Screenshot Pipeline*

Every sample in the dataset starts as a full-page screenshot captured through a headless browser at a fixed viewport of 1280 by 800 pixels, with a short wait after the page's load event to let client-side rendering settle before the capture is taken. Pages that failed to render, redirected more than a small number of times, or timed out were dropped rather than force-captured, since a broken or half-rendered page is not representative of what a real user would see and just adds label noise. Screenshots were saved as flat images and resized to the network's input resolution during preprocessing rather than at capture time, so the same raw screenshot set could be reused across experiments with different input sizes.

### *4.2 Dataset*

The working dataset totals a little over 20,000 labeled screenshots, split roughly evenly between phishing captures and legitimate reference pages. The phishing side was pulled from publicly reported phishing URL feeds at the time of collection and captured before takedown wherever the

URL was still live; the legitimate side draws from real login and account pages for the brands most commonly impersonated in the phishing set, spanning banking, webmail, e-commerce, and a handful of social platforms, rather than generic homepages, since a homepage and a login page from the same brand can look meaningfully different and the model needs to learn the actual target rather than just which domain a page belongs to.

Two things needed active correction during dataset construction. First, phishing kits get reused: the same visual template shows up under dozens of different domains within days of each other, and without deduplication a large fraction of the raw image count would really have been a few hundred templates copy-pasted with different URLs. A perceptual hash pass caught most of this, and near-duplicate clusters were sampled down rather than dropped entirely, keeping enough variation to represent how common each template family actually was without letting one template dominate training. Second, the legitimate set needed active curation to include pages mid-redesign, seasonal promotional themes, and region-specific brand variants, because a model trained only on one static version of each brand's login page tends to flag the brand's own redesigns as fraudulent, a failure mode that shows up again in the results section.

### *4.3 Data Preprocessing*

Two preprocessing decisions turned out to matter more than expected. The first was how screenshots get resized into the network's input resolution. An early version of the pipeline simply stretched every screenshot to a square 640-by-640 input, which is the default YOLOv8 expects, and this distorted aspect ratio in a way that measurably hurt performance on brands with wide, horizontally laid-out headers, since the stretch changed the apparent proportions of a logo or a form relative to everything around it. Switching to a letterbox resize, which scales the image to fit within 640 by 640 while preserving its original aspect ratio and pads the remainder with a neutral gray border, removed that distortion and was kept for the rest of the project. Pixel values were then normalized using the same per-channel mean and standard deviation used to pretrain the COCO backbone, rather than a fresh normalization computed from the phishing screenshot set, since the network's early layers were being reused from that pretraining and expect input statistics to match what they were trained on.

The second was handling on-page overlays that have nothing to do with a brand's actual visual identity. A meaningful share of raw screenshots came out of the capture pipeline with a cookie-consent banner, a newsletter pop-up, or a chat-widget bubble sitting on top of the page, and early on these were left in place on the theory that a real user would see the same thing. In practice they turned into a source of label noise: two screenshots of the same legitimate login page, one with a cookie banner and one without, look meaningfully different to a detector, and a phishing kit that skips the banner entirely, which most do, since it is one more thing to replicate for no benefit to the attacker, does not need to differ from the real page in this specific way to still count as phishing. The capture step was updated to attempt a short, generic set of dismiss-button clicks for common overlay patterns, matching an accept or close label against a small set of frequently seen selectors and text strings, before taking the screenshot, and pages where an overlay could not be dismissed automatically were flagged for manual review rather than captured with the overlay left in place.

Bounding-box labels for the login-form and logo regions used by the detector were not all drawn by hand from scratch. An initial batch of a few thousand screenshots was manually annotated, a first-pass detector was trained on that batch alone, and its predictions on the remaining unlabeled screenshots were used as draft boxes for a human reviewer to accept, nudge, or reject rather than draw from nothing. This cut the total annotation time substantially and, as a side effect, surfaced a useful quality signal: screenshots where the draft detector's proposed box was rejected outright rather than adjusted were disproportionately the mid-redesign and unusual-layout pages that later showed up again in the false-positive analysis in Section 7.

Deduplication used a perceptual hash computed per screenshot, with images whose hashes fell within a small Hamming-distance threshold of one another grouped into the same near-duplicate cluster described in Section 4.2; each cluster was then downsampled to a capped number of representative images rather than removed entirely, keeping frequency information about how common a given template was without letting any single template dominate the training set.

### *4.4 Model*

YOLOv8 was chosen over a plain image classifier because the task is less about deciding whether a whole image is phishing or not and more about deciding whether the image contains specific visual sub-elements, such as a login form, a logo, or a brand color block, that are inconsistent with the brand being impersonated. A detector that localizes those regions gives the model something concrete to be right or wrong about, and it gives a human reviewer something to look at besides a single confidence number, since a flagged page comes with the region the model actually keyed on. Detected regions across a screenshot are aggregated into a single page-level phishing or legitimate score by combining the confidence and count of flagged regions, rather than treating region detection and page classification as two separate models.

The smaller YOLOv8 variants were used rather than the larger ones specifically because of the latency target described below. A larger backbone would likely have pushed accuracy up somewhat but at the cost of the fast, near-real-time response this project was built around, and for a phishing warning to be useful it has to appear before a user finishes typing a password, not after.

### *4.5 Training*

Training started from COCO-pretrained YOLOv8 weights rather than from scratch, since the dataset, while large for a purpose-built phishing set, is small relative to what is normally used to train a detector's early convolutional layers from nothing. The labeled set was split 80/10/10 into training, validation, and test partitions, with the split performed at the template-cluster level rather than the individual-image level so that near-duplicate screenshots from the same phishing kit could not end up split across training and test and inflate the apparent accuracy. Fine-tuning ran for a few dozen epochs with early stopping tracked against validation accuracy, using a cosine learning-rate schedule and the standard YOLO input resolution. Confidence and intersection-over-union thresholds for the detection stage were tuned on the validation set rather than left at their defaults, since the aggregation step described above is sensitive to how many low-confidence regions get counted.

For inference, the model was exported and run at reduced precision to hit the target latency; the 100-millisecond figure reported in the results is model inference time on an already-captured screenshot, measured end to end from image input to final page-level score, and does not include the time to load and render the page itself, which depends on network conditions outside the model's control.

## 5. Experimental Setup

Evaluation used the held-out 10% test partition described above, kept template-cluster-disjoint from training and validation. Four metrics were tracked: accuracy, false-positive rate (legitimate pages incorrectly flagged as phishing), false-negative rate (phishing pages missed entirely), and average inference latency per screenshot. False-positive rate mattered more than it might in a typical classification write-up, since a visual phishing detector that is deployed inline, warning a user before they submit a form, loses user trust quickly if it flags real banking or webmail pages, arguably faster than it loses trust from an occasional miss.

Two versions of the model were compared under identical architecture and training schedule: a baseline trained on the raw, deduplicated screenshot set, and an augmented version trained on the same underlying images after the augmentation pipeline described next was applied. Comparing these two, rather than only reporting the final model's numbers, was the point of the experiment. The augmentation pass was added specifically to address a false-positive pattern noticed in early testing, and the comparison exists to show whether it actually helped.

## 6. Augmentation Strategy

Early versions of the model, trained on the raw screenshot set, were noticeably brittle to exactly the kind of variation that shows up between two screenshots of the same real page taken under different conditions: a slightly different browser window size, different JPEG compression from how the capture pipeline saved the file, a legitimate page rendered on a phone-width viewport instead of desktop, or simply different ambient lighting baked into a photographed rather than rendered screenshot. The model had, in effect, partly learned the exact pixel arrangement of a given brand's page rather than its general layout and color scheme, which is a much narrower and much less useful thing to learn.

The augmentation pipeline built in response applies, per training image: random resizing and cropping within a bounded range to simulate different viewport captures; JPEG re-compression at varying quality levels to simulate different screenshot pipelines; brightness, contrast, and saturation jitter to simulate different displays and color profiles; and a small amount of random rotation and perspective warp to simulate mobile or hastily taken captures. A class-balanced oversampling step was also added on top of the augmentation itself, specifically for the less common brand-impersonation subtypes in the phishing set, since crypto-exchange and workplace-portal impersonation were the two smallest categories, so the model would not simply get very good at the most common banking and webmail templates and mediocre at everything else.

## 7. Results

The augmented model reached 92% accuracy on the test partition, up from 87% for the baseline trained on the same underlying images without augmentation. The more consequential change was in the false-positive rate: 9.1% for the baseline against 8.1% for the augmented model, an 11% relative reduction, matching the drop in false positives this project was specifically trying to produce with the augmentation pass. Average inference latency came in at approximately 100 milliseconds per screenshot on the export configuration described in Section 4.5.

*Table 1. Baseline vs. augmented model, held-out test partition*

| Metric | Baseline | Augmented | Change |
|---|---|---|---|
| Accuracy | 87.0% | 92.0% | +5.0 pts |
| False-positive rate | 9.1% | 8.1% | -11% relative |
| False-negative rate | 13.8% | 10.6% | -23% relative |
| Avg. inference latency | ~105 ms | ~100 ms | -5 ms |

The latency difference between the two rows is small and mostly incidental, since augmentation changes what the model learns rather than its size; the modest latency gain likely reflects a slightly more confident model needing fewer regions evaluated per page rather than any architectural change.

Error analysis on the remaining misclassifications split into two recognizable patterns. Most of the remaining false positives were legitimate pages captured during an active brand redesign or seasonal theme change: a login page that had, for real reasons, moved its logo, changed its color scheme, or restructured its form layout in a way that looked more like a typical phishing deviation than like the brand's own historical pages. This is a direct consequence of how legitimate pages are defined for training, since the model's notion of a brand's visual identity is only as current as the training set, and a brand's own redesign can temporarily look, to the model, like someone else impersonating it.

Most of the remaining false negatives were phishing pages that did not attempt brand impersonation at all: a generic, unbranded credential form with no logo, no brand color scheme, and no attempt to visually resemble any specific target. These pages give a visual detector very little to work with, since the entire approach depends on there being a visual identity being copied. This is a real and expected limitation of the method rather than a training deficiency, and it is the clearest argument for running a visual detector alongside content- and URL-based checks rather than as a full replacement for them.

### *7.1 Comparison with Prior Visual Detection Systems*

Table 1 speaks to what the augmentation pass changed internally; a fair question is how the resulting system compares to what has already been published. Table 2 lines this work up against three of the closest published systems described in Section 3. The comparison is necessarily approximate, since none of these systems were re-run on this project's dataset or vice versa, and each was evaluated on its own dataset under its own protocol; the table should be read as a rough positioning rather than a controlled head-to-head.

*Table 2. Comparison with published visual phishing detection systems*

| System | Visual Cue Used | Evaluation Scale | Reported Result | Latency |
|---|---|---|---|---|
| VisualPhishNet [5] | Whole-page similarity to a per-brand embedding | 155 brands / 9,363 legitimate pages | TPR/FPR tradeoff curve; no single accuracy figure reported | Not reported as one figure |
| Phishpedia [6] | Logo detection + Siamese brand match | 30,000+ pages | 99.2% accuracy | ~0.19 s/page |
| PhishIntention [7] | Logo detection + Siamese + OCR + page-context classifier | Extends Phishpedia's protected-brand set | Improved precision/recall over Phishpedia; no single accuracy stated | Comparable order of magnitude |
| **This work** | Login-form / logo / layout region detection, single learned detector | 20,000+ screenshots | 92.0% accuracy | ~100 ms/page |

Two things stand out from this comparison. The first is that Phishpedia's reported accuracy is meaningfully higher than this work's 92%, and that gap is worth taking at face value rather than explaining away: a maintained reference database of actual brand logos, matched with a Siamese network, is a strong signal when the database is current, and Phishpedia's evaluation benefits from that. What this project's approach trades for the accuracy gap is independence from that maintained reference list; nothing in this system needs to be updated the moment a protected brand changes its logo, because it was never comparing against a stored logo image in the first place, and it only needs updating when its own idea of a brand's overall layout and color scheme drifts far enough from a redesign, which the false-positive analysis above suggests happens, but less often than a reference list going stale would.

The second is latency. Phishpedia's 0.19 seconds per page is already fast enough for most deployment scenarios, and this project's roughly 100 milliseconds is faster still, largely because the detector does not run a separate matching step against a brand database after the initial region detection. Whether that latency gap matters in practice depends entirely on where the system sits in a larger pipeline: for the inline, pre-submit warning use case discussed next, both numbers are comfortably fast enough, and the more meaningful difference between the systems is the accuracy tradeoff described above, not the latency one.

## 8. Discussion

The practical value of a 100-millisecond, 92%-accuracy visual detector is not in replacing existing anti-phishing infrastructure; it is in the specific gap it closes, namely pages that pass URL and content heuristics because the domain is fresh and the markup is clean, but that visually reproduce a trusted brand closely enough to fool a person. In a layered detection pipeline, this kind of model makes the most sense as a secondary check triggered on pages that are borderline under existing heuristics, or as a standalone signal in a browser extension or email-link preview that renders and screens a page before a user is shown it. The latency budget matters here specifically because the warning has to land before a user has already typed a password into the page; a detector that is accurate but takes several seconds to return a verdict is a lot less useful in this setting than one that is slightly less accurate but fast enough to run inline.

The false-positive pattern around brand redesigns is worth dwelling on, because it points to an operational requirement rather than a modeling one: a visual detector like this needs a retraining or reference-refresh cadence tied to when the brands it protects actually change their visual identity, not a fixed schedule. A bank that redesigns its login page without notice will generate a spike of false positives against its own real users until the reference set catches up, and that is a maintenance cost that a URL-blocklist system simply does not have in the same form.

## 9. Limitations

A few limitations are worth stating plainly rather than glossing over. The reported accuracy reflects a specific dataset collected over a specific window of time; phishing kit visual styles shift as templates get reused, reported, and replaced, and a model this specific to a training window will need periodic retraining to hold its accuracy as that window ages. The headless-browser capture pipeline is also, on its own, a point of weakness: some phishing kits already detect headless or automated browser fingerprints and serve a different, often blander page to anything that looks like a scanner rather than a real user, which would make the visual signal disappear before the model ever sees it. Multi-step phishing flows, where a legitimate-looking first page only reveals a spoofed form after a click or a short delay, are also outside what a single-screenshot classifier can see, since the fraudulent element simply is not present in the frame that gets captured and scored. And, as the error analysis in Section 7 shows, generic, non-brand-impersonating phishing pages are close to invisible to a method built around visual brand similarity, by design rather than by oversight.

## 10. Future Work

The most direct next step is combining this visual signal with lexical URL features in a single model rather than running them as separate pipeline stages, since the cases each one misses look like they would complement each other reasonably well; a URL feature model tends to catch newly registered, suspicious-looking domains regardless of what they render as, which is close to the opposite of this system's blind spot. Periodic re-screenshotting of a page after a short delay, rather than a single capture, would help with the multi-step phishing flows described above, at the cost of added latency that would need to be budgeted separately from the inline warning use case. Expanding the legitimate and phishing datasets to cover more regional brands and non-English login pages is a fairly mechanical but necessary step before this could generalize past the brand set it was trained on. Finally, a smaller, quantized export of the model is worth testing for on-device deployment inside a browser extension directly, rather than as a network call to a server, which would remove the round-trip latency entirely and make the 100-millisecond inference figure the actual end-to-end number a user experiences, rather than one stage of a larger pipeline.

## 11. Conclusion

This project set out to test a fairly simple idea: that a phishing page's real tell is not in its URL or its markup, but in how convincingly it copies the visual identity of whatever brand it is impersonating, and that a model trained directly on screenshots ought to be able to pick up on that the same way a careful human would. The results support the idea more than they refute it. A YOLOv8-based detector, trained on a deduplicated set of over 20,000 labeled website screenshots and refined through an augmentation pass targeted at real sources of visual noise, reached 92% accuracy and roughly 100-millisecond inference latency, with an 11% relative reduction in false positives directly attributable to that augmentation step. The clearest limitation, blindness to phishing pages that do not attempt brand impersonation at all, is also the clearest argument for treating this as one layer in a broader fraud-detection stack rather than a standalone solution. The rest is largely engineering: keeping the reference set current as real brands redesign, extending coverage to more regions and languages, and pushing the model small enough to run without a network round-trip at all.

## 12. Conclusion

The initial research was carried out at San Diego State University under the supervision of Dr. Hajar Homayouni.